\documentclass[runningheads]{llncs}

\usepackage{eccv}

\usepackage{eccvabbrv}

\usepackage{url}
\usepackage{graphicx}
\usepackage{booktabs} 
\usepackage{bbding}
\usepackage{pifont}
\usepackage{wasysym}
\usepackage{amssymb}

\usepackage{subcaption}

\usepackage[accsupp]{axessibility}  

\usepackage[pagebackref,breaklinks,colorlinks,citecolor=eccvblue]{hyperref}

\usepackage{orcidlink}
\usepackage{makecell}

\begin{document}

\title{Social-Mamba: Socially-Aware Trajectory Forecasting with State-Space Models} 

\titlerunning{Social-Mamba}

\author{Po-Chien Luan\inst{1}\orcidlink{0000-0003-3788-093X} \and
Wuyang Li\inst{1}\orcidlink{0000-0002-7338-9251} \and
Yang Gao\inst{1}\orcidlink{0000-0002-3695-9155} \and
Alexandre Alahi \inst{1}\orcidlink{0000-0002-5004-1498}}

\authorrunning{P.-C. Luan et al.}

\institute{EPFL, Switzerland, \\
\email{firstname.lastname@epfl.ch}\\}

\maketitle

\begin{abstract}

Human trajectory forecasting is crucial for safe navigation in crowded environments, requiring models that balance accuracy with computational efficiency. Efficiently modeling social interactions is key to performance in dense crowds. Yet, most recent methods rely on attention mechanisms, which are effective at capturing complex dependencies, but incur quadratic computational costs that scale poorly with the growing number of neighbors. Recently, Selective State-Space Models have provided a linear-time alternative; however, their inherently sequential design is misaligned with the unstructured and dynamic nature of social interactions. To address this challenge, we propose Social-Mamba, a forecasting architecture that reformulates social interactions as structured sequential processes. At its core is the Cycle Mamba block, a novel module that enables continuous bidirectional information flow. Social-Mamba organizes agents on an egocentric grid and introduces social triplet factorization, which decomposes interactions into temporal, egocentric, and goal-centric scans. These are dynamically integrated through a learnable social gate and global scan to generate accurate and efficient trajectory predictions. Extensive experiments on five trajectory forecasting benchmarks show that Social-Mamba achieves state-of-the-art accuracy while offering superior parameter efficiency and computational scalability. Furthermore, embedding Social-Mamba into a flow-matching framework further enhances both accuracy and efficiency, establishing it as a flexible and robust foundation for future trajectory forecasting research. The code is publicly available:  \href{https://github.com/vita-epfl/Social-Mamba}
{\color{magenta}https://github.com/vita-epfl/Social-Mamba}.
\end{abstract}
\section{Introduction}

Human trajectory forecasting is essential for safe, real-time decision-making in applications ranging from autonomous robotics to sports analytics~\cite{gao2024multi,li2025voxdet}. The key challenge lies in balancing predictive accuracy and computational efficiency in dynamic, multi-agent environments, where inaccurate or slow predictions \cite{li2025stable} can lead to unsafe outcomes \cite{gao2026deformable}.

A central factor in accurate forecasting is the ability to model complex social interactions. Early deep learning methods relied on LSTMs to encode agent histories, combined with social pooling to aggregate neighbor influences \cite{alahi2016social}. Graph Neural Networks introduced more structure by representing agents as nodes and enabling explicit message passing \cite{salzmann2020trajectron++}. More recently, Transformers have become the dominant paradigm, using self-attention \cite{vaswani2017attention} to capture global all-to-all interactions and delivering significant improvements~\cite{yuan2021agentformer,girgis2021latent,saadatnejad2023social,gao2025social}. However, the \textbf{quadratic cost of attention} \cite{vaswani2017attention} with respect to the number of agents, along with more parameters, makes it prohibitively expensive in crowded scenes.

Recently, the research community has turned to more efficient architectures, State Space Models (SSMs), particularly Mamba~\cite{gu2023mamba}, as a highly promising solution to relieve this computational barrier. With their linear-time complexity and proven success in modeling long-range dependencies in other domains, SSMs are theoretically well-suited for processing long temporal sequences in trajectory data. Several pioneering works have attempted to leverage these advantages \cite{capellera2025unified, xu2024sports, huang2025trajectory, zhang2024decoupling}. However, these attempts have revealed a fundamental conflict between the native design of SSMs and the nature of multi-agent dynamics. 

Directly applying SSMs to model social interactions reveals two critical, unresolved challenges. First, there is a \textbf{fundamental structural mismatch}: SSMs are designed for ordered, one-dimensional sequences, whereas social interactions are unstructured in 2D or 3D space. Forcing agents into an arbitrary sequence is not only unnatural but also risks destroying crucial spatial information and imposing an artificial hierarchy. Second, a significant \textbf{information flow mismatch exists}. Standard SSMs process information unidirectionally, yet social interactions are often holistic and bidirectional. The intent behind an agent's early action is often clarified only by subsequent events, a retrospective context that a purely forward-pass model cannot capture.

\begin{figure}[t]
\begin{center}
\includegraphics[width=1.0\linewidth]{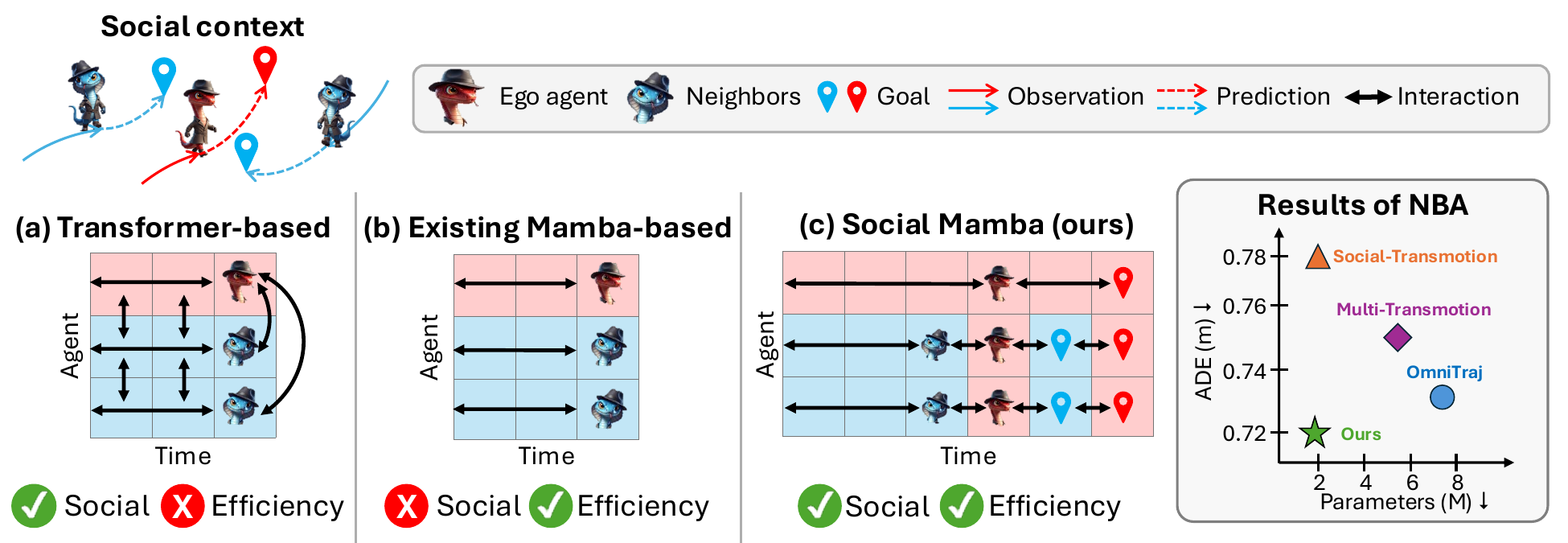}
\caption{\textbf{Comparison of interaction modeling approaches.} (a) Transformers capture all-pair interactions but incur quadratic complexity.  (b) Prior Mamba-based models are restricted to temporal-only modeling. (c) Our Social-Mamba integrates ego-agent context to enable efficient social interaction modeling, achieving state-of-the-art accuracy with up to 75\% fewer parameters.}
\label{fig:pull}
\end{center}
\end{figure}

In addressing these issues, prior work has relied on workarounds rather than fundamental solutions. Some models restrict Mamba to modeling only the temporal history of individual agents, sidestepping the social modeling problem entirely \cite{capellera2025unified, xu2024sports}. Others alter Mamba's core mechanism, reformulating it as an attention-like operator to handle unstructured data \cite{huang2025trajectory}, thereby sacrificing the unique properties of the original SSM architecture. Recently, MambaPTP \cite{zhang2025mambaptp} pioneered the use of a pure Mamba for trajectory prediction. However, it restricts social interaction modeling to the decoding stage and applies generic sequential scanning to neighbor embeddings, leaving the structural mismatch between 1D SSMs and unstructured social scenes unaddressed. Consequently, no prior work has successfully used an SSM in its native form for social interaction reasoning, leaving its full potential for trajectory forecasting unrealized.

To bridge this gap, we propose \textbf{Social-Mamba}, a novel architecture that adapts the sequential nature of SSMs to effectively model unstructured social dynamics. Our key idea is to integrate social interaction into conventional temporal scanning by imposing a meaningful structure on the scene and decomposing interactions into selective sequential scans. First, we preprocess agents into an ego-centric social grid that organizes neighbors into a preferred structure from the ego agent's perspective, thereby resolving the ordering problem. Next, we design a \textbf{social triplet factorization}, which replaces one complex scan with three distinct sequential scans: a temporal scan for individual agent histories, an ego-centric scan for neighbor influence on the ego’s current state, and a goal-centric scan for how neighbors affect the ego’s path toward its goal. Interaction is enabled by inserting the ego’s state and goal tokens into neighbor sequences before scanning, allowing different interactions to be processed in a temporal manner. We show the main comparison in \cref{fig:pull}.
Finally, we perform \textbf{social fusion}, where context-dependent weights are learned for each interaction in the triplet, followed by a global scan to aggregate the information.

At the heart of these modules is the \textbf{Cycle Mamba} (CM) block, a novel bidirectional SSM designed for richer contextual understanding. In social scenarios, early actions are often reinterpreted by later events, and neighbor influence on the ego agent is inherently bidirectional. Standard bidirectional models struggle with this \cite{zhang2024motion,capellera2025unified,xu2024sports, capellera2025jointdiff}, as their forward and backward passes remain disconnected. CM resolves this by concatenating the forward sequence with its reverse, thereby ensuring uninterrupted hidden-state flow. This allows the backward pass to be explicitly conditioned on the forward context, mirroring human social reasoning. Furthermore, the single-pass design can reduce the number of parameters by half, thereby improving memory efficiency.

In summary, Social-Mamba is the first trajectory forecasting architecture built entirely on Mamba, focusing on bridging the gap between sequential SSMs and unstructured social interactions. By introducing the CM, we further strengthen bidirectional continuity and save parameters. Extensive experiments demonstrate that Social-Mamba achieves state-of-the-art accuracy on five benchmark datasets while offering superior efficiency and computational scalability.

\section{Related Work}

\subsection{Human Trajectory Forecasting}

Research in human trajectory forecasting has advanced along two main axes: \emph{modeling social interactions} and \emph{capturing the uncertainty of future paths}.  For interaction modeling, early work introduced \emph{social pooling} mechanisms that aggregated neighboring context in an unstructured way \cite{alahi2016social,kothari2021human}. To impose more explicit structure, GNNs emerged as a dominant paradigm, representing agents as nodes and enabling message passing for structured reasoning \cite{huang2019stgat, li2020evolvegraph,salzmann2020trajectron++, xu2022groupnet, mohamed2020social}. More recently, \emph{attention-based methods} have provided even greater flexibility, dynamically weighting all agents to capture global, context-dependent influences \cite{messaoud2020attention, giuliari2021transformer, yuan2021agentformer, Girgis2021LatentPrediction, luan2025unified}.  Orthogonal to interaction modeling is the challenge of multimodality, as multiple futures are often plausible. Addressing this requires generative approaches, which have evolved from GANs \cite{gupta2018social, sadeghian2019sophie, kosaraju2019social} and VAEs \cite{schmerling2018multimodal, ivanovic2020multimodal, xu2022socialvae} to more powerful diffusion-based techniques \cite{gu2022trajdiffusion, mao2023leapfrog, bae2024singulartrajectory} and, most recently, flow matching models \cite{fu2025moflow} and interactive models \cite{sun2025interactive}, which generate higher-quality and more consistent trajectory distributions.

\subsection{Mamba for Trajectory Forecasting}

Recent works have begun to adopt the Mamba architecture to improve temporal modeling in trajectory prediction. For example, U2Diff \cite{capellera2025unified} replaces the Transformer encoder with a bidirectional Mamba, thereby enabling richer temporal dynamics without positional encodings. Similarly, Sports-Traj \cite{xu2024sports} incorporates a bidirectional temporal Mamba within its Transformer encoder to capture temporal dependencies in sports data. In the autonomous driving domain, Tamba \cite{huang2025trajectory} leverages Mamba to redesign the encoder–decoder architecture, replacing self-attention with a more efficient linear-time alternative. Despite these advances, a key limitation remains: existing approaches restrict Mamba to temporal modeling and do not explicitly capture social interactions between agents. This leaves the potential of SSMs for multi-agent reasoning largely unexplored.

\section{Preliminaries}

\label{subsec:ssm}
 \textbf{Selective State Space Models}. Classical SSMs~\cite{kalman1960new} describe sequential data using hidden dynamics governed by linear systems of differential equations. Recently, structured SSMs such as S4~\cite{gu2021efficiently} showed that these models can be adapted for deep learning: by parameterizing the system matrices and discretizing with efficient rules such as zero-order hold (ZOH), S4 achieves linear-time sequence modeling while retaining strong long-range memory. This makes S4 a competitive alternative to quadratic-cost Transformers~\cite{vaswani2017attention}. However, S4 learns a time-invariant system, where parameters remain fixed across the input. To increase flexibility, Mamba~\cite{gu2023mamba} introduces selectivity: system parameters are conditioned on the input, allowing the model to adapt its dynamics over time. Formally, given input $x_t$, hidden state $h_t$, and output $y_t$, the continuous dynamics are:

\begin{equation}
h'(t) = \mathbf{A}(t) h(t) + \mathbf{B}(t) x(t), 
\quad y(t) = \mathbf{C}(t) h(t),
\end{equation}
where $\mathbf{A}(t)$ is the state matrix,  $\mathbf{B}(t)$ is the input matrix, and $\mathbf{C}(t)$ is the output matrix, which can be discretized as:

\begin{equation}
h_t = \mathbf{\bar{A}}(t) h_{t-1} + \mathbf{\bar{B}}(t) x_t, 
\quad y_t = \mathbf{C}(t) h_t,
\end{equation}
with $\mathbf{\bar{A}}(t)$ and $\mathbf{\bar{B}}(t)$ computed via ZOH. Selective SSMs differ from S4 by replacing the fixed parameters $(\Delta, \mathbf{B}, \mathbf{C})$ with input-dependent mappings:
\begin{equation}
\Delta(t) = \tau_\Delta(\Delta + s_\Delta(x_t)), \quad
\mathbf{B}(t) = s_B(x_t), \quad
\mathbf{C}(t) = s_C(x_t),
\end{equation}
where $\tau_\Delta$ is a softplus function and $s_\Delta, s_B, s_C$ are learned projections. This selective mechanism enables the model to modulate temporal-level information flows, effectively bridging the efficiency of SSMs with the adaptability of Transformers.

\label{subsec:trajectory}

\noindent\textbf{Ego-Centric human trajectory prediction.}  
The goal of ego-centric trajectory forecasting is to predict the future path of a specific ego agent, denoted as $e$, by considering its own motion history as well as the social context provided by all neighboring agents. Formally, consider $N$ agents observed for $T_{\text{obs}}$ time steps, where the trajectory of agent $i$ is $X_i = (x_i^1, x_i^2, \dots, x_i^{T_{\text{obs}}}), \; x_i^t \in \mathbb{R}^2$. The set of observed trajectories for all agents is $\mathbf{X} = \{X_e, X_1, X_2, \dots, X_N\}$. 
The task is to learn a mapping $f$ that takes the histories of all agents and predicts the future motion of only the ego agent $e$ over the next $T_{\text{pred}}$ time steps: $f: \mathbf{X} \mapsto \hat{Y}_e$, where the predicted trajectory $\hat{Y}_e = (\hat{y}_e^{T_{\text{obs}}+1}, \dots, \hat{y}_e^{T_{\text{obs}}+T_{\text{pred}}}), \; \hat{y}_e^t \in \mathbb{R}^2$.

\section{Methodology}

We first introduce the Cycle Mamba block, the core module that enables continuous bidirectional information flow within our framework. Building on this foundation, we then present the complete Social-Mamba architecture, which integrates structured representations of social interactions with novel scanning strategies for trajectory forecasting.

\subsection{Cycle Mamba Block}
\label{subsec:smiling_mamba}
\begin{figure}[t]
  \centering
  \includegraphics[width=1.0\linewidth]{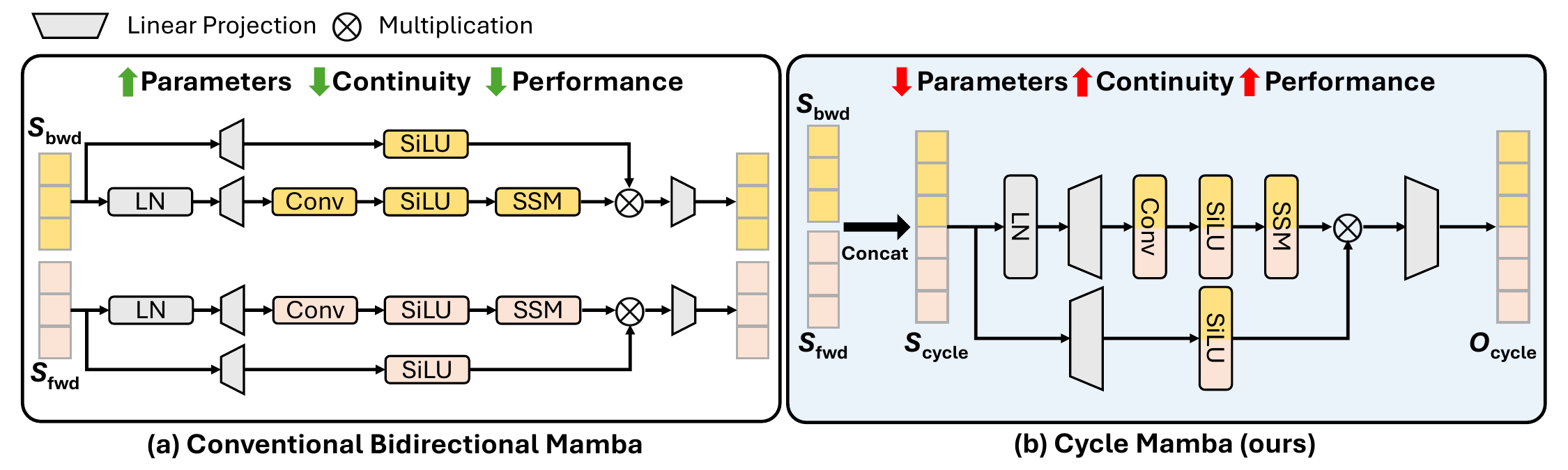}
  \caption{\textbf{Comparison of Bidirectional Mamba architectures.} (a) A conventional bidirectional Mamba uses two isolated passes, suffering from a contextual disconnect as information is only fused at the output layer. (b) CM performs a single, continuous scan. This ensures state continuity by having the forward pass directly initialize the backward pass, enabling a more integrated fusion with significantly fewer parameters.}
   \label{fig:smile}
\end{figure}

A fundamental challenge in modeling sequential data, such as trajectories, is capturing context from both past and future elements. Conventional recurrent architectures typically address this through contextual isolation: two independent models process the forward and backward sequences separately, and their outputs are combined only at a final late fusion stage. While functional, this separation prevents the model from learning the seamless, continuous dependencies that exist between the two directions.

To overcome this limitation, we propose the CM, an architecture that unifies bidirectional processing into a single, continuous scan, as shown in \cref{fig:smile}. By creating a \textbf{cycle} sequence, the model's internal state propagates naturally from the backward context directly into the forward pass, enabling a more deeply integrated state-level fusion. The mechanism is formalized as follows:

    \noindent \textbf{Sequence construction}: Given an input sequence of $L$ vectors, $S_{\text{fwd}} = (s_1, \allowbreak s_2, \dots, s_L)$, where $s_t \in \mathbb{R}^D$, we first construct its reverse, $S_{\text{bwd}} = (s_L,\allowbreak s_{L-1}, \dots, s_1)$. The cycle sequence, $S_{\text{cycle}} \in \mathbb{R}^{2L \times D}$, is formed by their concatenation:
\begin{equation}
    S_{\text{cycle}} = [S_{\text{bwd}} ; S_{\text{fwd}}] = (s_L, \dots, s_1, s_1, \dots, s_L).
\end{equation}
   \textbf{Continuous state-space scan}: A single Mamba model, defined by its state-space parameters $(\bar{A}, \bar{B}, C, D)$, processes this unified sequence. The hidden state $h_k \in \mathbb{R}^N$ evolves over the $2L$ timesteps according to the recurrence:
\begin{equation}
    h_k = \bar{A} h_{k-1} + \bar{B} u_k, \quad y_k = C h_k + D u_k,
\end{equation}
where $u_k$ is the $k$-th vector of $S_{\text{cycle}}$. The complete output is $O_{\text{cycle}} = (o_1, \dots, o_{2L})$.

    \noindent \textbf{Output reconstruction}: The output sequence $O_{\text{cycle}}$ is deconstructed back into its backward and forward components. Let $O_{\text{bwd}} = (o_1, \dots, o_L)$ and $O_{\text{fwd}} = (o_{L+1}, \dots, o_{2L})$. The final output $O \in \mathbb{R}^{L \times D'}$ is obtained by aligning and merging these components, for example through element-wise addition:
\begin{equation}
    O = O_{\text{fwd}} + \text{flip}(O_{\text{bwd}}).
\end{equation}

\noindent The novelty of CM lies in its information flow. In a standard bidirectional model, the forward pass begins with a zero-initialized state $h_0^f = 0$, making it ignorant of the future. In our approach, the forward pass (processing $x_1$ at timestep $k=L+1$) is initialized with the state $h_L$. This state is the final hidden state of the backward pass over $(x_L, \dots, x_1)$. By unrolling the recurrence, we see that $h_L$ is a comprehensive summary of the entire reversed sequence:
\begin{equation}
    h_L = \sum_{i=1}^{L} \bar{A}^{L-i} \bar{B} x_{L-i+1}.
\end{equation}
Therefore, the very first step of the forward pass, computing the state for $x_1$, is:
\begin{equation} \label{eq:state_propagation}
    h_{L+1} = \bar{A} h_L + \bar{B} x_1.
\end{equation}
Equation~\ref{eq:state_propagation} demonstrates that the forward pass is \textbf{explicitly conditioned on a compressed representation of the entire future context}. Instead of relying on a late fusion of two independent analyses, CM performs an \textbf{integrated, state-level fusion} where the forward representation is a direct, causal function of the backward one. This continuous propagation of the hidden state allows for a more sophisticated and cohesive model of bidirectional dependencies, which is especially critical for complex sequential tasks like trajectory forecasting.

\subsection{Social-Mamba}

\begin{figure}[t]
  \includegraphics[width=1.0\linewidth]{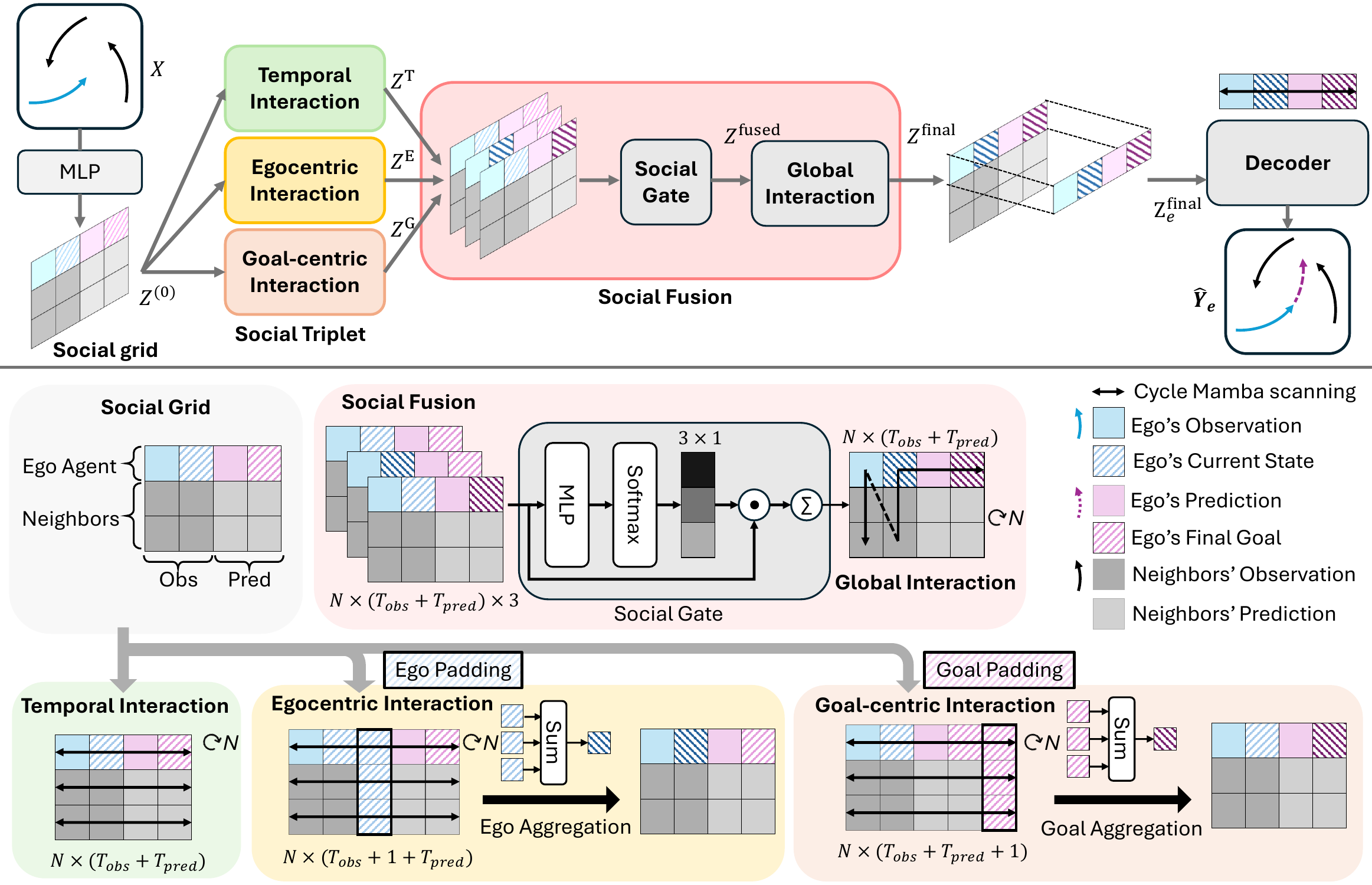}
  \caption{\textbf{Overview of the Social-Mamba framework}. The model first establishes an ego-centric view by creating a sorted social grid. This grid is then processed by three parallel interaction modules—temporal, egocentric, and goal-centric—which use our Cycle Mamba blocks to capture different facets of social influence. A dynamic gating network fuses these representations, and a final scan across agents captures global interactions to produce the ego-centric trajectory prediction.}
   \label{fig:social_mamba}
\end{figure}

Social-Mamba is designed to learn hierarchical social representations through a multi-stage process. It begins by structuring the scene from an ego-centric viewpoint and then applies a series of specialized interaction modules (see \cref{fig:social_mamba}).

\noindent\textbf{Social grid preparation.} The set of observed trajectories is defined as $\mathbf{X} = \{X_e, X_1, \dots, X_N\}$, where $X_i = (x_i^1, \dots, x_i^{T_{\text{obs}}})$ and $x_i^t \in \mathbb{R}^2$. To capture the relevant social context for a selected ego agent $e$, we define a local social circle by retaining only the agents located within a specified spatial radius (e.g., 10 m) at time $T_{\text{obs}}$. The trajectories of these $D$ filtered agents are then used to construct the input tensor $\mathbf{S} \in \mathbb{R}^{D \times (T_{\text{obs}} + T_{\text{pred}}) \times 2}$. The future coordinates are initialized with zero vectors. This tensor is then encoded into a higher-dimensional representation, our initial social grid $\mathbf{Z}^{(0)} \in \mathbb{R}^{N \times T \times D}$, where $T = T_{obs} + T_{pred}$ and $D$ is the model dimension:
\begin{equation}
    \mathbf{Z}^{(0)} = \text{MLP}(\mathbf{S}).
\end{equation}

\noindent\textbf{Social triplet interaction.} We factorize social interactions into three parallel streams, where $\mathbf{Z}^{(0)}_i \in \mathbb{R}^{T \times D}$ is the sequence for agent $i$. We use $\text{CM}(\cdot)$ to denote the Cycle Mamba block.

\begin{itemize}
    \item \textbf{Temporal interaction:} Each agent's sequence is processed independently to capture individual motion dynamics:
    \begin{equation}
        \mathbf{Z}^{T}_i = \text{CM}(\mathbf{Z}^{(0)}_i).
    \end{equation}
    \item 
     \textbf{Egocentric interaction:} To model the influence of the ego agent's current state, its token at $T_{obs}$, denoted $z_{e, T_{obs}}^{(0)}$, is inserted into each neighbor's sequence. Let $\mathbf{Z'}_j$ be the augmented sequence for neighbor $j$:
    \begin{equation}
        \mathbf{Z'}_j = [z_{j,1}^{(0)}, ..., z_{j,T_{obs}}^{(0)}, z_{e,T_{obs}}^{(0)}, z_{j,T_{obs}+1}^{(0)}, ..., z_{j,T}^{(0)}],
    \end{equation}
    where $\mathbf{Z'}_j \in \mathbb{R}^{(T+1) \times D}$, and we use $\text{CM}(\cdot)$ to scan along the time axis:
    \begin{equation}
\mathbf{Z'}^{E}j = \text{CM}(\mathbf{Z'}j), \quad z_{e, T_{obs}}^{E} = \sum_{t=T_{obs}}^{T_{obs}+1} w_{t} \cdot z'_{e, t} + \sum_{j=1}^{N} w_j \cdot z'_{j,T_{obs+1},}
    \end{equation}

    where $\mathbf{Z}^{'E}_j \in \mathbb{R}^{(T+1) \times D}$, and $\mathbf{Z}^{'E}_j =[z'_{j,1}, ..., z'_{j,T_{obs}}, z'_{e,T_{obs}}, z'_{j,T_{obs}+1}, ..., z'_{j,T}]$.  The output is processed by learnable weighted sums and abandons the padding tokens to maintain the same shape that $ \mathbf{Z}^{E}_j \in \mathbb{R}^{T \times D}$.
     \item 
     \textbf{Goal-centric interaction:} Similarly, to model the influence of the ego's goal, its token at time $T$, $z_{e,T}^{(0)}$, is appended to each neighbor's sequence $\mathbf{Z''}_j$: $\mathbf{Z''}_j = [z_{j,1}^{(0)}, ..., z_{j,T}^{(0)}, z_{e,T}^{(0)}]$. We use $\text{CM}(\cdot)$ to scan along the time axis:
    \begin{equation}
        \mathbf{Z}^{''G}_j = \text{CM}(\mathbf{Z''}_j), \quad z_{e, T}^{G} =\sum_{t=T}^{T+1} w_{t} \cdot z''_{e, t} + \sum_{j=0}^{N} w_j \cdot z''_{j,T+1},
    \end{equation}

where $\mathbf{Z}^{''G}_j \in \mathbb{R}^{(T+1) \times D}$, and $\mathbf{Z}^{'E}_j =[z''_{j,1}, ..., z''_{j,T_{obs}}, z''_{e,T_{obs}}, z''_{j,T_{obs}+1}, ..., z''_{j,T}]$. We use learnable weights to sum the goal tokens to merge information. To maintain the same shape, we remove the padding tokens that $ \mathbf{Z}^{G}_j \in \mathbb{R}^{T \times D}$.
\end{itemize}
\textbf{Social fusion and decoding.} The three representations, $\mathbf{Z}^{T}, \mathbf{Z}^{E}, \mathbf{Z}^{G}$, are fused using a dynamic gating mechanism. First, they are concatenated, and a gating network computes weights for each interaction type:
\begin{equation}
    \mathbf{W} = \text{Softmax}(\text{MLP}(\text{Concat}(\mathbf{Z}^{T}, \mathbf{Z}^{E}, \mathbf{Z}^{G}))),
\end{equation}
where $\mathbf{W} = \{w_T, w_E, w_G\}$. The fused representation $\mathbf{Z}^{\text{fused}}$ is a weighted sum:
\begin{equation}
    \mathbf{Z}^{\text{fused}} = w_T \odot \mathbf{Z}^{T} + w_E \odot \mathbf{Z}^{E} + w_G \odot \mathbf{Z}^{G}.
\end{equation}
To model global interactions, we apply a Mamba scan along the agent axis (the first dimension of $\mathbf{Z}^{\text{fused}}$). Finally, the socially-aware representation for the ego agent, $\mathbf{Z}^{\text{final}}_e$, is isolated and passed to a final decoder network, a simple bidirectional Mamba and $K$ MLP projection heads, to predict the future trajectory:
\begin{equation}
    \hat{Y}_e = \text{Decoder}(\mathbf{Z}^{\text{final}}_e),
\end{equation}
where $\hat{Y}_e \in \mathbb{R}^{K \times T_{pred} \times 2}$ is the final predicted $K$ trajectories of the ego agent.

\subsection{Loss}
To account for the inherent multimodality of human motion, our model predicts $K$ possible future trajectories. We train the model using a best-of-K loss strategy. Let $\mathbf{\hat{Y}}_e = \{\hat{Y}_{e,1}, ..., \hat{Y}_{e,K}\}$ be the set of $K$ predicted trajectories for the ego agent $e$, and let $Y_e$ be the corresponding ground truth trajectory. The loss for this sample is the Mean Squared Error (MSE), denoted as,
\begin{equation}
    \mathcal{L}_e = \min_{k \in \{1,...,K\}} \frac{1}{T_{pred}} \sum_{t=T_{obs}+1}^{T_{obs}+T_{pred}} ||\hat{y}_{e,k}^t - y_e^t||_2^2.
\end{equation}
The final training objective is the average of this loss over all samples in a batch. This encourages the model to generate at least one plausible and accurate future path among its $K$ hypotheses.

\section{Experiments}
\subsection{Datasets}
\textbf{NBA.} The NBA dataset contains the trajectories of all 10 players and the ball during professional basketball games \cite{Nba2016}. The constant presence of a fixed number of agents makes it a unique testbed for analyzing complex group dynamics. We follow the standard setup of predicting 20 future frames (4.0s) from 10 past frames (2.0s). To ensure comprehensive evaluation, we adopt three widely used splits: (i) the full split (NBA-Full) introduced in \cite{gao2024multi}, which leverages all data in NBA-LED \cite{mao2023leapfrog}, and (ii) scenario-specific splits focusing on rebounding and scoring plays \cite{xu2022socialvae}, (iii) the mini split (NBA-LED). Our main experiments are conducted on NBA-Full, with additional evaluations on the other splits

\noindent\textbf{Stanford Drone Dataset (SDD).} SDD captures real-world pedestrian dynamics from a bird’s-eye view across a university campus \cite{robicquet2016learning}. Its diverse trajectories and crowded interactions make it well-suited for evaluating social forecasting models. Following prior work \cite{xu2022socialvae}, we use 8 observed frames (3.2s) to predict the subsequent 12 frames (4.8s). All trajectories are processed in meters.

\noindent\textbf{JackRabbot Dataset and Benchmark (JRDB).} JRDB is a large-scale, egocentric dataset recorded from a mobile social robot navigating both indoor and outdoor environments \cite{martin2021jrdb}. It features diverse social interactions between humans, robots, and static obstacles, making it particularly challenging. Following the standard protocol in \cite{fangneuralized}, we predict 12 future frames (4.8s) based on 9 observed frames (3.6s), with trajectories sampled at 2.5 Hz.

\subsection{Metrics}
We evaluate our model using standard metrics for multimodal trajectory forecasting, where the model predicts $K$ possible future trajectories.

\noindent\textbf{Minimum Average Displacement Error (minADE$_{K}$):} This is the average L2 distance between the ground-truth trajectory and the closest of the $K$ predicted trajectories, calculated over all timesteps in the prediction horizon.

\noindent\textbf{Minimum Final Displacement Error (minFDE$_{K}$):} the L2 distance between the final ground-truth position and the endpoint of the closest predicted trajectory among the $K$ candidates.  For simplicity, we refer to these metrics as ADE and FDE throughout the paper. Unless otherwise noted, we set $K=20$ and report all errors in meters.

\subsection{Quantitative Results}

\begin{table}[h]
\centering
\caption{\textbf{Comparison with baseline models on NBA-Full.} Models marked with * are pretrained on large-scale trajectory datasets. Best results are highlighted in bold.}
\label{tab:NBA1}
\resizebox{0.75\linewidth}{!}{%
\begin{tabular}{lcccc}
\toprule
 \textbf{Method} & \textbf{ADE {$\downarrow$}} & \textbf{FDE {$\downarrow$}} & \textbf{Parameters (M) {$\downarrow$}}  & \textbf{GFLOPs {$\downarrow$}}\\
\midrule
Social-Transmotion \cite{saadatnejad2023social} & 0.78 & 1.01 & 2.0 & 0.87\\
Multi-Transmotion* \cite{gao2024multi} &  0.75 & 0.97 & 5.7 & 0.87\\
OmniTraj* \cite{gao2025omnitraj}&  0.73 & 0.94 & 7.5 & 1.45\\
Social-Mamba (ours) &  \textbf{0.72} & \textbf{0.92} & \textbf{1.9} & \textbf{0.66}\\
\bottomrule
\end{tabular}
}
\end{table}

\noindent\textbf{NBA-Full.} Table \ref{tab:NBA1} reports results on the NBA dataset. Social-Mamba consistently outperforms all baselines in both ADE and FDE. Notably, several strong competitors such as Multi-Transmotion and OmniTraj rely on pretraining with large-scale external datasets before fine-tuning. In contrast, Social-Mamba achieves superior accuracy by training solely from scratch on the official dataset, highlighting both its data efficiency and the effectiveness of its architecture.

\noindent\textbf{NBA Scoring and Rebounding.} As shown in \cref{tab:NBA2}, Social-Mamba also achieves state-of-the-art results on the specialized Scoring and Rebounding splits. The model shows a particularly strong advantage in the Scoring scenario, where coordinated, goal-directed behavior is critical. In this setting, Social-Mamba reduces ADE by 8.2\% and FDE by 7.3\% compared to prior work, highlighting its effectiveness at modeling structured group interactions.

\noindent\textbf{SDD.} We report all results in meters and compare against baselines that follow the same protocol. Reporting in meters provides a standardized and meaningful comparison, as pixel-based metrics are influenced by camera parameters and resolution, whereas metric distances are absolute. For completeness, we also provide pixel-based results in the appendix. In \cref{tab:SDD}, Social-Mamba achieves performance comparable to the current state-of-the-art on this challenging benchmark.

\noindent\textbf{JRDB.} In \cref{tab:JRDB}, Social-Mamba consistently outperforms prior methods across different prediction horizons, demonstrating both stability and robustness. In particular, our method achieves substantial gains, improving ADE by up to 13\% and FDE by 8.7\% over the previous state-of-the-art.

\begin{table}[t!]
\centering
\begin{minipage}{0.45\textwidth}
    \centering
    \caption{Comparison with baseline on NBA Rebounding and Scoring. 
    }
    \label{tab:NBA2}
    \resizebox{\textwidth}{!}{
    \begin{tabular}{lcc}
        \toprule
        \textbf{Method} & \textbf{Rebounding} & \textbf{Scoring} \\
        \midrule
        Trajectron++ \cite{salzmann2020trajectron++} & 0.98/1.93 & 0.73/1.46 \\
        BiTrap \cite{yao2021bitrap} & 0.83/1.72 & 0.74/1.49 \\
        SGNet-ED \cite{wang2022stepwise} & 0.78/1.55 & 0.58/1.30 \\
        Social-VAE \cite{xu2022socialvae} & 0.72/1.37 & 0.64/1.17 \\
        RNLS \& CLLS \cite{qiu2025adapting} & 0.65/1.20 & 0.61/1.09 \\
        Social-Mamba (ours) & \textbf{0.63/1.18} & \textbf{0.56/1.01} \\
        \bottomrule
    \end{tabular}}
\end{minipage}
\hfill
\begin{minipage}{0.35\textwidth}
    \centering
    \caption{Comparison with baseline models on SDD.}
    \label{tab:SDD}
    \resizebox{\textwidth}{!}{
    \begin{tabular}{lc}
        \toprule
        \textbf{Method} & \textbf{ADE/FDE{$\downarrow$}} \\
        \midrule
        Trajectron++ \cite{salzmann2020trajectron++} & 0.34/0.58 \\
        BiTrap \cite{yao2021bitrap} & 0.32/0.57 \\
        SGNet-ED \cite{wang2022stepwise} & 0.33/0.58 \\
        Social-VAE \cite{xu2022socialvae} & 0.27/0.39 \\
        NMRF \cite{fangneuralized} & \textbf{0.25}/0.39 \\
        Social-Mamba (ours) & \textbf{0.25/0.38} \\
        \bottomrule
    \end{tabular}}
\end{minipage}
\end{table}

\begin{table}[t!]
\centering
\begin{minipage}{0.56\textwidth}
    \centering
    \caption{Comparison with baseline models on JRDB across different horizons. 
    }
    \label{tab:JRDB}
    \resizebox{\textwidth}{!}{
    \begin{tabular}{lcccc}
        \toprule
        \textbf{Method} & \textbf{1.2s} & \textbf{2.4s} & \textbf{3.6s} & \textbf{4.8s (total)} \\
        \midrule
        LED \cite{mao2023leapfrog} & 0.05/0.07 & 0.09/0.14 & 0.14/0.21 & 0.18/0.28 \\ 
        NMRF \cite{fangneuralized} & \textbf{0.04/0.05} & 0.08/0.11 & 0.11/0.17 & 0.15/0.23 \\
        Social-Mamba (ours) & \textbf{0.04/0.05} & \textbf{0.07/0.10} & \textbf{0.10/0.15} & \textbf{0.13/0.21} \\
        \bottomrule
    \end{tabular}}
\end{minipage}
\hfill
\begin{minipage}{0.40\textwidth}
    \centering
    \caption{Quantitative results of MoFlow using different social encoders on NBA-LED. 
    }
    \label{tab:moflow}
    \resizebox{\textwidth}{!}{
    \begin{tabular}{lcc}
        \toprule
        \textbf{MoFlow encoder} & \textbf{ADE/FDE {$\downarrow$}} & \textbf{Parameters (M) {$\downarrow$}} \\
        \midrule
        Transformer & 0.71/0.87 & 1.3 \\ 
        Social-Mamba & \textbf{0.70/0.85} & \textbf{0.5} \\
        \bottomrule
    \end{tabular}}
\end{minipage}
\end{table}

\subsection{Efficiency}
 A central motivation of our work is to improve the efficiency of trajectory forecasting models. All experiments were conducted on a single NVIDIA A100 GPU, with a batch size of 1 during inference and 128 during training. As shown in \cref{tab:NBA1}, Social-Mamba achieves state-of-the-art accuracy on the NBA dataset while requiring significantly fewer parameters and lower computational cost (GFLOPs) than competing methods. Compared to the prior state-of-the-art, our model reduces parameters by 75\% and GFLOPs by 54\%. To further validate the practicality of our approach, we provide additional evaluation of computational efficiency, including inference time, memory usage of models, and GPU cost during training, following similar settings in \cite{zhang2025mambaptp}.

\noindent \textbf{Interaction module analysis.} We first examine the efficiency of the proposed Cycle Mamba block compared to a standard Bidirectional Mamba and the MHSA. As shown in \cref{tab:infer_mem_interaction}, Cycle Mamba achieves the lowest memory footprint (7.3 MB) and the fastest inference time (3.4 ms). Notably, it outperforms the standard Bidirectional Mamba in speed, likely due to its continuous scan design, which avoids the overhead of managing two independent, disconnected passes. All modules maintain low resource consumption, confirming that the core building blocks of Social-Mamba are lightweight. In training, Cycle Mamba can be trained with the lowest GPU cost under the same batch size, demonstrating that our model has a lower hardware requirement during training.

\noindent\textbf{Model-level comparison.} We further extend this evaluation to the full model level, comparing Social-Mamba against recent Transformer-based baselines: Social-Transmotion \cite{saadatnejad2023social} and Multi-Transmotion \cite{gao2024multi}. The results in \cref{tab:infer_mem_model} demonstrate Social-Mamba's superior scalability. While maintaining state-of-the-art accuracy, Social-Mamba operates with minimal resource consumption, requiring just 7.3 MB of memory and 3.4 ms inference time. This represents a significant efficiency gain over complex architectures such as Multi-Transmotion, which require over $7.3\text{ ms}$ per inference. It is worth noting that Social-Transmotion achieves a slightly faster inference time ($1.8\text{ ms}$); this is attributed to the Transformer's inherent advantage in parallel computation for short sequences, a behavior consistent with observations in \cite{zhang2025mambaptp}. Nevertheless, Social-Mamba offers the most favorable trade-off between high accuracy (as detailed in the main results) and a low memory footprint, making it well-suited for deployment in real-time systems.

\begin{table*}[t!]
    \centering
    
    \begin{minipage}{0.55\linewidth}
        \centering
        \caption{Efficiency comparison of different interaction modules.}
        \label{tab:infer_mem_interaction}
        \resizebox{0.95\linewidth}{!}{
        \begin{tabular}{lccc}
        \toprule
        \textbf{Model} & \textbf{\makecell{Inference \\ time (ms)}} & \textbf{\makecell{Model \\ memory (MB)}} & \textbf{\makecell{Training \\ memory (MB)}}\\
        \midrule
        MHSA & 4.4 & 8.7 & 14356 \\
        Bidirectional Mamba & 5.1 & 9.1 & 9422\\
        Cycle Mamba (ours) & \textbf{3.4} & \textbf{7.3} & \textbf{9375} \\
        \bottomrule
        \end{tabular}}
    \end{minipage}
    \hfill 
    \begin{minipage}{0.41\linewidth}
        \centering
        \caption{Efficiency comparison of different forecasting models.}
        \label{tab:infer_mem_model}
        \resizebox{0.95\linewidth}{!}{
        \begin{tabular}{lcc}
        \toprule
        \textbf{Model} & \textbf{\makecell{Inference \\ time (ms)}} & \textbf{\makecell{Model \\ memory (MB)}} \\
        \midrule
        Social-Transmotion & \textbf{1.8} & 7.6 \\
        Multi-Transmotion & 7.3 & 21.8 \\
        Social-Mamba (ours) & 3.4 & \textbf{7.3} \\
        \bottomrule
        \end{tabular}}
    \end{minipage}
    
\end{table*}

\noindent\textbf{Flexibility.} This inherent efficiency also makes the architecture highly flexible. To demonstrate this, we integrated Social-Mamba as a drop-in replacement for the Transformer-based encoder in a state-of-the-art flow-matching framework \cite{fu2025moflow}. The implementation details are in the appendix. Results in \cref{tab:moflow} show that this integration not only improves accuracy but does so with an encoder that is 2.3× smaller, highlighting Social-Mamba’s practicality as a lightweight and general-purpose social interaction module.

\subsection{Qualitative Results}

\begin{figure}[t!]
\centering
  \includegraphics[width=1.0\linewidth]{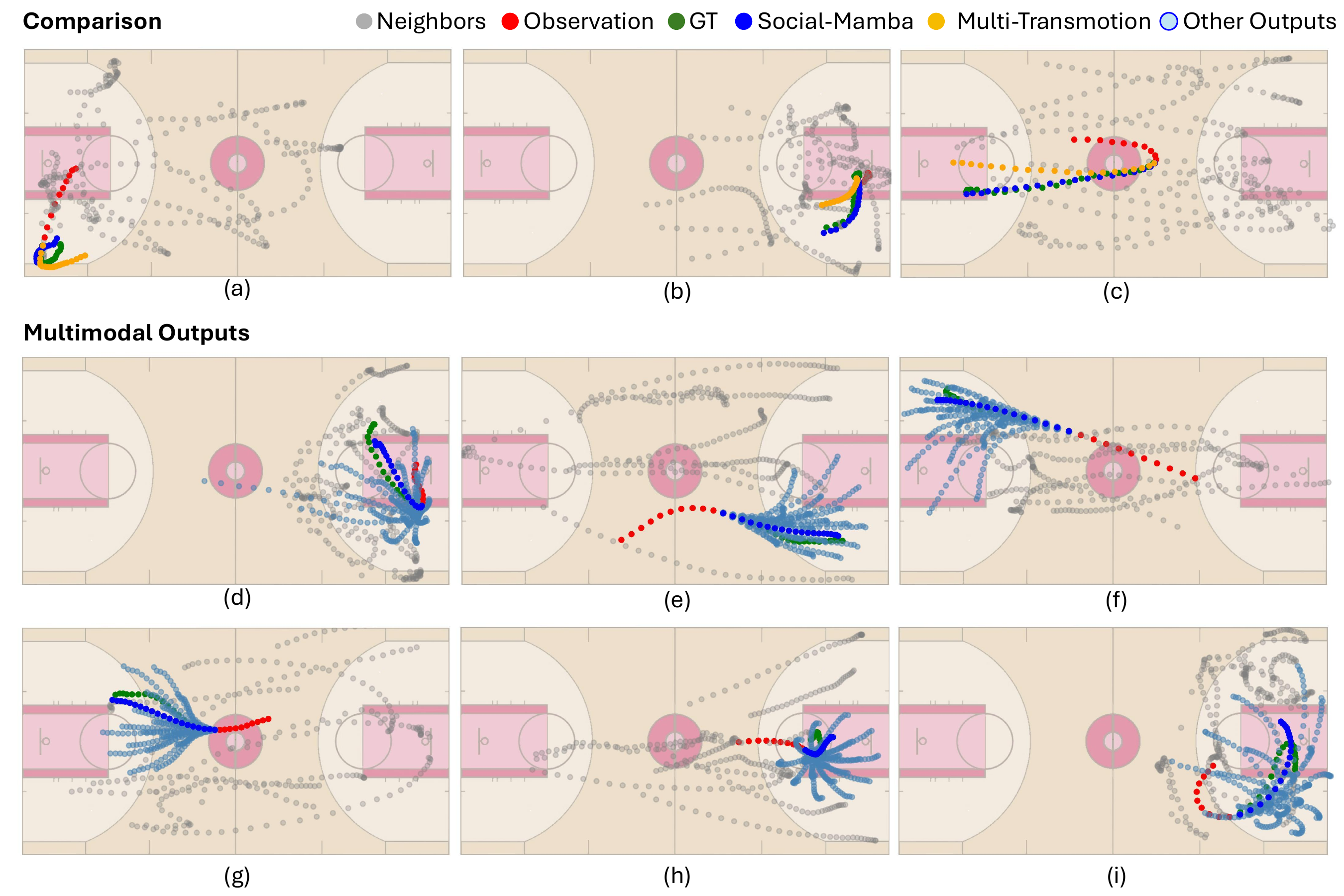}
  \caption{\textbf{Qualitative results of the NBA-Full}. (a)-(c): Comparison with Multi-Transmotion. (d)-(i): Multimodal outputs.}
   \label{fig:Qualitative}
   \centering
\end{figure}

To further illustrate performance, we visualize predictions on NBA-Full in comparison with Multi-Transmotion in \cref{fig:Qualitative} (a)-(c). Social-Mamba consistently produces more accurate trajectories across diverse scenarios, including abrupt turns, direction changes, and running to the other side of the field. (d)-(i) visualizes the complete set of output modalities to assess prediction diversity. Social-Mamba generates trajectories that span plausible futures with realistic directions and velocities, showcasing its ability to capture multimodal behaviors. In (h), the play occurs in the paint during an attack, leading to more diverse predictions. We also note occasional boundary violations, as shown in (i). 

\begin{figure}[t!]
\centering
  \includegraphics[width=0.8\linewidth]{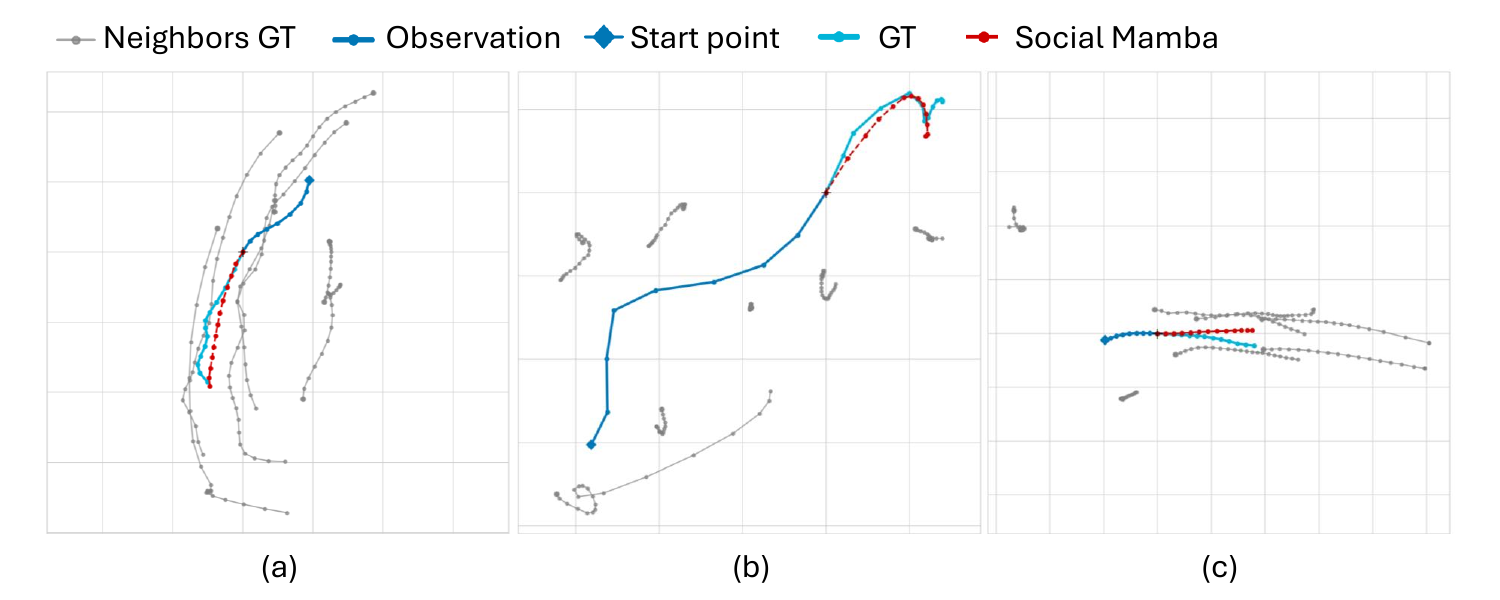}
    \caption{\textbf{Additional qualitative results on JRDB.} (a) Crowded environment. (b) Non-linear trajectory. (c) Failure case.}
   \label{fig:Qualitative_extra_jrdb}
\centering
\end{figure}

In JRDB (\cref{fig:Qualitative_extra_jrdb}), we visualize crowded campus scenes.  (a) shows Social-Mamba handling dense crowds, (b) demonstrates accurate predictions of highly non-linear behaviors, and (c) highlights a case with small deviations, where the prediction still avoids potential collisions. Additional results are in the appendix.
\subsection{Ablation Studies}
\noindent\textbf{Impact of interaction modules}. We analyze the contribution of each interaction module. To preserve essential temporal and spatial reasoning, the temporal and global scans are kept as the baseline. As shown in \cref{tab:alation_modules}, removing either the goal-centric or ego-centric scan results in worse performance, while combining both yields the best performance, highlighting their complementary roles.

\begin{table}[t!]
\centering
\begin{minipage}{0.37\textwidth}
    \centering
    \caption{\textbf{Ablation of the interaction modules}.}
    \label{tab:alation_modules}
    \resizebox{\textwidth}{!}{
    \begin{tabular}{lcccc}
        \toprule
        \textbf{Temporal} & \textbf{Ego} & \textbf{Goal} & \textbf{Global} & \textbf{ADE/FDE {$\downarrow$}} \\
        \midrule
        \checkmark & \ding{55} & \ding{55} & \checkmark & 0.735/0.939 \\
        \checkmark & \checkmark & \ding{55} & \checkmark & 0.729/0.928 \\
        \checkmark & \ding{55} & \checkmark & \checkmark & 0.727/0.928 \\
        \checkmark & \checkmark & \checkmark & \checkmark & \textbf{0.719/0.919} \\
        \bottomrule
    \end{tabular}}
\end{minipage}
\hfill
\begin{minipage}{0.55\textwidth}
    \centering
    \caption{\textbf{Comparison with different interaction blocks}.}
    \label{tab:ablation_block}
    \resizebox{\textwidth}{!}{
    \begin{tabular}{lccc}
        \toprule
        \textbf{Interaction block} & \textbf{ADE/FDE {$\downarrow$}} & \textbf{Parameters (M) {$\downarrow$}}  &\textbf{GFLOPs {$\downarrow$}} \\
        \midrule
        MHSA & 0.741/0.947 & 2.3 & 0.80 \\
        Bidirectional Mamba & 0.741/0.948 & 2.4 & \textbf{0.66} \\
        Cycle Mamba (ours)& \textbf{0.719/0.919} & \textbf{1.9} & \textbf{0.66} \\
        \bottomrule
    \end{tabular}}
\end{minipage}
\end{table}

\noindent\textbf{Architecture.} We compare our CM block against other interaction mechanisms, including a standard bidirectional Mamba \cite{zhang2024motion, xu2024sports, capellera2025unified} and MHSA \cite{vaswani2017attention}. As shown in \cref{tab:ablation_block}, CM achieves the best performance with the fewest parameters, demonstrating its effectiveness as an efficient building block.

We further analyze our architectural choices for the social triplet. Specifically, we compare our parallel design with a sequential variant where temporal, egocentric, and goal-centric interactions are applied in order. As shown in \cref{tab:ablation_seq}, the parallel design achieves superior performance. We attribute this to our parallel fusion gate, which enables the model to dynamically adjust the contribution of each interaction type. In contrast, a sequential design risks information loss, as features from earlier interactions may be overwritten by subsequent ones. We also investigate how to fuse the outputs of the social triplet (\cref{tab:ablation_fusion}). We compare our learnable softmax gating mechanism with a simple additive fusion that directly sums the three interaction terms without weights. The results demonstrate that incorporating learnable weights provides greater flexibility across scenarios. For instance, when the ego agent moves independently without nearby interactions, the model can automatically downweight the influence of social context, while in crowded scenes it can emphasize neighbor interactions more strongly.

\noindent\textbf{Decoder structure.} We compare our Mamba-based decoder to a conventional MLP decoder \cite{saadatnejad2023social, gao2024multi} as shown in \cref{tab:ablation_decoder}. The results suggest that the Mamba decoder is more effective at generating high-quality trajectories.
\begin{table}[t!]
\centering
\begin{minipage}{0.3\textwidth}
    \centering
    \caption{\textbf{Comparison with sequential social triplet.}}
    \label{tab:ablation_seq}
    \resizebox{0.9\textwidth}{!}{
    \begin{tabular}{lc}
        \toprule
        \textbf{Social triplet} & \textbf{ADE/FDE {$\downarrow$}} \\
        \midrule
        Sequential & 0.728/0.929 \\
        Parallel & \textbf{0.719/0.919} \\
        \bottomrule
    \end{tabular}}
\end{minipage}
\hfill
\begin{minipage}{0.33\textwidth}
    \centering
    \caption{\textbf{Comparison of fusion strategies for the social triplet.}}
    \label{tab:ablation_fusion}
    \resizebox{0.9\textwidth}{!}{
    \begin{tabular}{lc}
        \toprule
        \textbf{Social fusion} & \textbf{ADE/FDE {$\downarrow$}} \\
        \midrule
        Addition & 0.744/0.954 \\
        Learnable weights & \textbf{0.719/0.919} \\
        \bottomrule
    \end{tabular}}
\end{minipage}
\hfill
\begin{minipage}{0.25\textwidth}
    \centering
    \caption{\textbf{Comparison with MLP decoder.}}
    \label{tab:ablation_decoder}
    \resizebox{0.85\textwidth}{!}{
    \begin{tabular}{lc}
        \toprule
        \textbf{Decoder} & \textbf{ADE/FDE {$\downarrow$}} \\
        \midrule
        MLP & 0.730/0.931 \\
        Mamba & \textbf{0.719/0.919} \\
        \bottomrule
   \end{tabular}}
\end{minipage}
\end{table}

\section{Conclusion}
We introduced Social-Mamba, the first forecasting architecture built entirely on the Mamba framework, to address the fundamental challenge of applying sequential SSMs to unstructured social dynamics. Our model bridges this gap by structuring the scene with an ego-centric social grid and decomposing interactions via a social triplet factorization, powered by our novel bidirectional Cycle Mamba block. Across five benchmark datasets, Social-Mamba achieves new state-of-the-art performance with significantly greater computational efficiency and demonstrated flexibility, confirmed by its successful integration into a flow matching framework.

\section*{Acknowledgment}

This work was funded by Honda R\&D Co., Ltd, SwissAI, and Sportradar. We would like to express our gratitude to Guillem Capellera and Reyhaneh Hosseininejad for their insightful feedback.
%
%
\bibliographystyle{splncs04}
\bibliography{main}
\newpage
\section{Appendix}
\noindent
The appendix provides additional details and analyses to complement the main paper. We begin with implementation details, including training configurations and architectural settings. Next, we present further experimental results, including pixel-based evaluation on SDD and performance on the deterministic TrajNet++ benchmark. To highlight the flexibility of our approach, we describe the integration of Social-Mamba into the MoFlow framework. Additional qualitative results are provided for NBA-Full, including successful and failure cases. 

\subsection{Implementation Details}
Our model was trained on a single NVIDIA A100 GPU. The feature dimension of the social grid representation is set to 128. Within each Mamba block, the SSM state dimension is 16, and the convolution kernel size is 4. We trained the model for 100 epochs using the ADAM optimizer and a step-based learning rate schedule to improve convergence.

\subsection{Additional Experimental Results}
\label{sdd_additional}
For completeness, we report results on SDD in pixel-based coordinates in \cref{tab:sdd_p}. Social-Mamba achieves performance comparable to the state-of-the-art method NMRF. However, we argue that pixel-based evaluation is unreliable: the real-world distance represented by a pixel varies with camera perspective and calibration. This introduces inconsistencies that hinder fair comparison.
We advocate reporting in meters as a standardized metric. As shown in \cref{tab:SDD}, Social-Mamba slightly outperforms NMRF under this evaluation. The discrepancy, the ranking of top methods changes between metrics, highlights the inconsistency of pixel-based evaluation and the reliability of real-world coordinates.

To further validate our model, we also evaluate Social-Mamba on a deterministic benchmark TrajNet++ \cite{kothari2021human}. As shown in \cref{tab:trajnet}, Social-Mamba performs strongly under this deterministic setup, confirming its robustness across different evaluation protocols.

\begin{table}[t!]
\caption{\textbf{Comparison with baseline models on SDD.} The numbers are in pixels. Underlines denote the second best.}
\label{tab:sdd_p}
\centering
\resizebox{0.55\textwidth}{!}{
\begin{tabular}{lcc}
\toprule
Method & \textbf{Venue}  & \textbf{ADE/FDE {$\downarrow$}} \\
\midrule
Trajectron++ \cite{salzmann2020trajectron++} & ECCV'20 & 10.00/17.15 \\ 
BiTrap \cite{yao2021bitrap} & RAL'21 & 9.09/16.31 \\
MID \cite{gu2022stochastic}& CVPR'22 & 9.08/17.04 \\
MemoNet \cite{xu2022remember}& CVPR'22 & 8.56/12.66 \\
SGNet-ED \cite{wang2022stepwise} & RAL'22 & 9.69/17.01 \\
Social-VAE \cite{xu2022socialvae} & ECCV'22 & 8.10/11.72\\
LED \cite{mao2023leapfrog}& CVPR'23 & 8.48/11.66 \\
TUTR \cite{shi2023trajectory} & ICCV'23 & 7.76/12.69 \\
ET-HighGraph \cite{kim2024higher} & CVPR'24 & 7.81/\textbf{11.09} \\
MoFlow-IMLE \cite{fu2025moflow}& CVPR'25 & 7.85/12.86 \\
UniTraj \cite{xu2024sports}& ICLR'25 & 8.68/12.78\\
NMRF \cite{fangneuralized} & ICLR'25 & \textbf{7.20}/ \underline{11.29}\\
Social-Mamba  (ours)& - & \underline{7.54}/11.74 \\

\bottomrule
\end{tabular}}
\end{table}

\begin{table}[t!]
\caption{\textbf{Comparison with baseline models on Trajnet++.} The ADE and FDE are deterministic.}
\label{tab:trajnet}
\centering

\resizebox{0.45\textwidth}{!}{
\begin{tabular}{lc}
\toprule
\textbf{Method} & \textbf{ADE/FDE {$\downarrow$}} \\
\midrule
DagNet  \cite{monti2021dag} & 0.66/1.44 \\
Trajectron++ \cite{salzmann2020trajectron++} & 0.55/1.16 \\ 
AutoBots \cite{girgis2021latent} & 0.54/1.12\\
Multi-Transmotion \cite{gao2024multi} & 0.54/1.13 \\
Social-Mamba  (ours) & \textbf{0.53/1.10} \\

\bottomrule
\end{tabular}}
\end{table}

\subsection{Social Mamba with Flow Matching}
\label{appendix_moflow}
\begin{figure}[t]
  \includegraphics[width=0.98\linewidth]{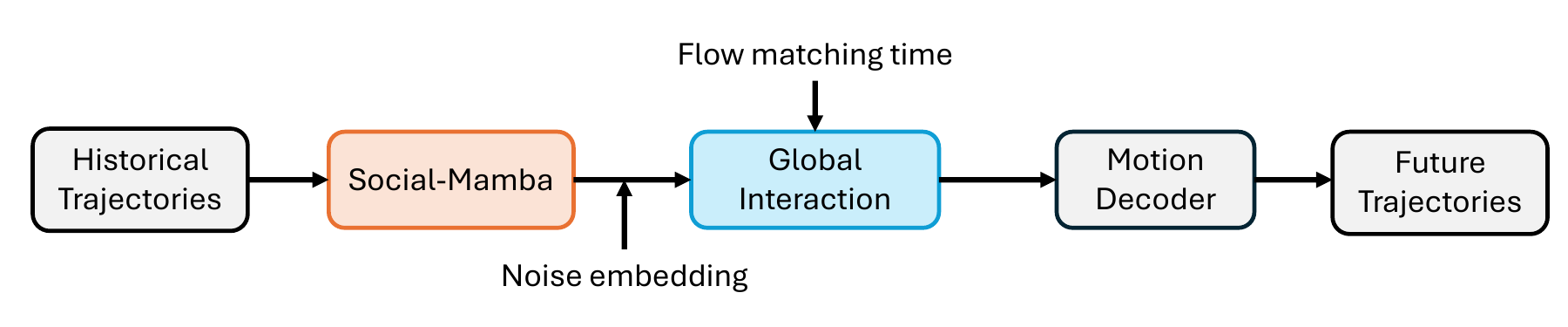}
  \caption{\textbf{Pipeline of MoFlow with Social Mamba}. We utilize Social-Mamba to model context and social-temporal interactions for MoFlow.}
   \label{fig:moflow-social-mamba}
\end{figure}

We integrated Social-Mamba into the MoFlow framework by replacing its Transformer-based context encoder with our model, and substituting the spatial-temporal Transformer with our global interaction scanning module. The motion decoder of MoFlow is left unchanged. The complete pipeline is illustrated in \cref{fig:moflow-social-mamba}.

\subsection{Additional Qualitative Results}
\label{appendix_qua}
\label{appendix_qua}

\begin{figure}[t!]
  \includegraphics[width=1.0\linewidth]{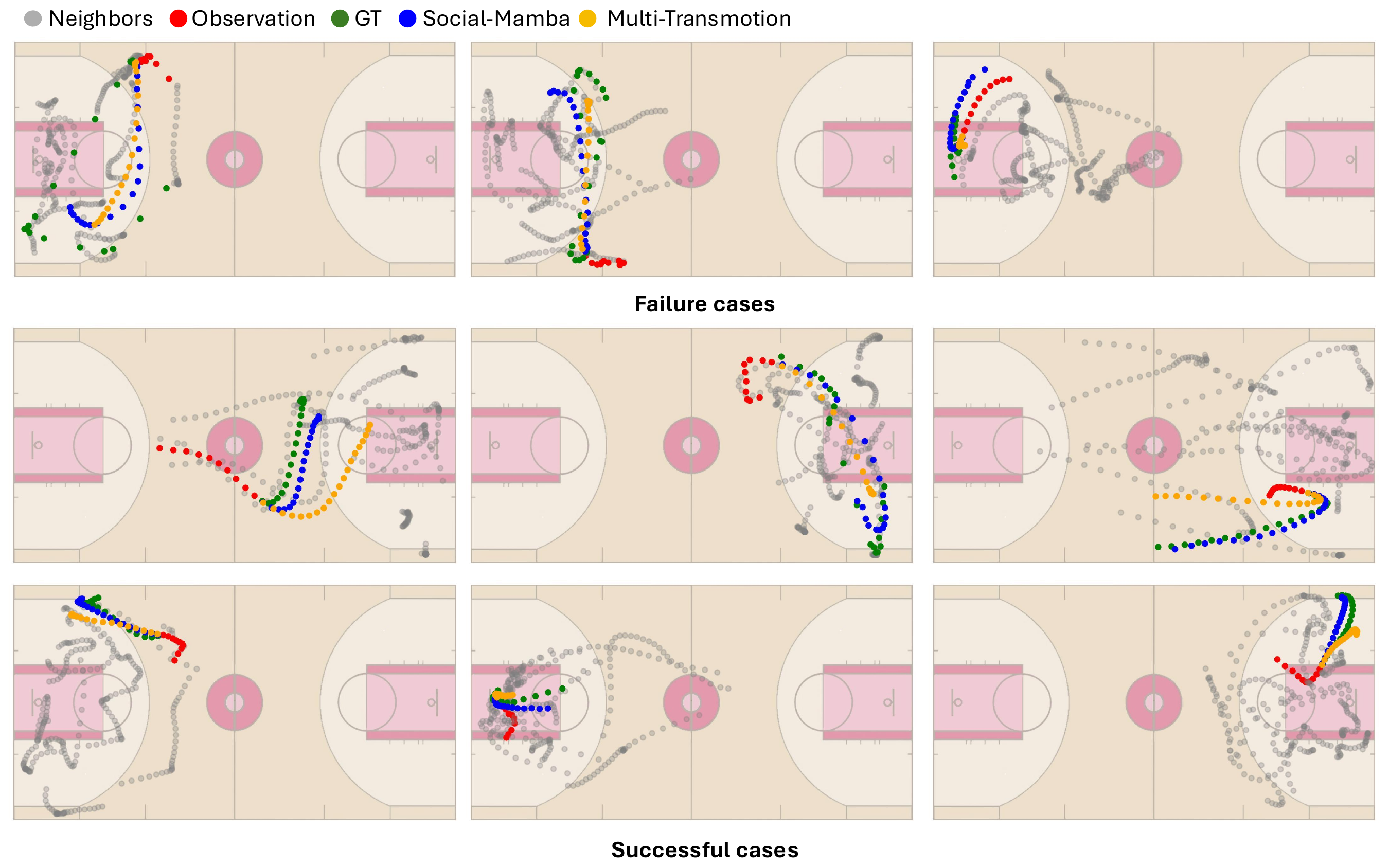}
  \caption{\textbf{Additional qualitative results of the NBA-Full}. Row 1 represents the failure cases. Rows 2 and 3 demonstrate successful cases.}
   \label{fig:Qualitative_extra}
\end{figure}

Figure \ref{fig:Qualitative_extra} presents extra visualizations on NBA-Full. We include failure cases, where errors often occur during ball passes—events that induce abrupt velocity changes and irregular trajectories. Sudden turns also remain challenging. Nonetheless, Social-Mamba generally provides more accurate predictions than Multi-Transmotion.

\subsection{In memory of Kobe Bryant (1978-2020)}
The development of our Social-Mamba model on the NBA dataset offered a moment for reflection. Kobe Bryant's words, "The most important thing is to try and inspire people so that they can be great in whatever they want to do," resonate with the broader goal of scientific research: to push boundaries and empower others to achieve greatness. This spirit is embodied in his "Mamba Mentality"—a relentless dedication to strategic preparation and an intuitive understanding of complex dynamics. As the first to apply a Mamba-based architecture to this dataset, we found inspiration in this parallel. This section is a small tribute to a legend whose mentality continues to inspire the pursuit of excellence in all fields.

\end{document}